\documentclass[letterpaper]{article} 
\usepackage{aaai25}  
\usepackage{times}  
\usepackage{helvet}  
\usepackage{courier}  
\usepackage[hyphens]{url}  
\usepackage{graphicx} 
\usepackage{natbib}  
\usepackage{caption} 
\usepackage{algorithm}
\usepackage{algorithmic}

\usepackage{newfloat}
\usepackage{listings}
\usepackage{newfloat}
\usepackage{listings}
\usepackage{color}
\usepackage{tabularray}
\usepackage{multirow}
\usepackage{amsmath} 
\usepackage{amsfonts} 
\usepackage{tabularx}
\usepackage{subfig}
\usepackage{array}
\usepackage{fontawesome}
\usepackage{colortbl}

\DeclareCaptionStyle{ruled}{labelfont=normalfont,labelsep=colon,strut=off} 
\floatstyle{ruled}
\newfloat{listing}{tb}{lst}{}
\floatname{listing}{Listing}
\title{MSE-Adapter: A Lightweight Plugin Endowing LLMs with the Capability to Perform Multimodal Sentiment Analysis and Emotion Recognition}
\author{
	Yang Yang, Xunde Dong\thanks{Xunde Dong is the corresponding author.}, Yupeng Qiang
}
\affiliations{
    South China University of Technology, School of Automation Science and Engineering\\
	
	China, Guangzhou\\
	audxd@scut.edu.cn
	
}

\usepackage{bibentry}

\begin{document}
	
	\maketitle
	
	\begin{abstract}
		Current Multimodal Sentiment Analysis (MSA) and Emotion Recognition in Conversations (ERC) methods based on pre-trained language models exhibit two primary limitations:
		1) Once trained for MSA and ERC tasks, these pre-trained language models lose their original generalized capabilities. 2) They demand considerable computational resources.  As the size of pre-trained language models continues to grow, training larger multimodal sentiment analysis models using previous approaches could result in unnecessary computational cost.  In response to this challenge, we propose \textbf{M}ultimodal \textbf{S}entiment Analysis and \textbf{E}motion Recognition \textbf{Adapter} (MSE-Adapter), a lightweight and adaptable plugin. This plugin enables a large language model (LLM) to carry out MSA or ERC tasks with minimal computational overhead (only introduces approximately 2.6M to 2.8M trainable parameters upon the 6/7B models),  while preserving the intrinsic capabilities of the LLM. In the MSE-Adapter, the Text-Guide-Mixer (TGM) module is introduced to establish explicit connections between non-textual and textual modalities through the Hadamard product. This allows non-textual modalities to better align with textual modalities at the feature level, promoting the generation of higher-quality pseudo tokens. Extensive experiments were conducted on four public English and Chinese datasets using consumer-grade GPUs and open-source LLMs (Qwen-1.8B, ChatGLM3-6B-base, and LLaMA2-7B) as the backbone. The results demonstrate the effectiveness of the proposed plugin.

	\end{abstract}
\begin{links}
	\link{Code}{https://github.com/AZYoung233/MSE-Adapter}
\end{links}
	%
	
	\section{Introduction}
	Making artificial intelligence (AI) comprehend human sentiment is a significant issue in the development of AI. Multimodal Sentiment Analysis (MSA) and Emotion Recognition in Conversations (ERC) have received widespread attention from the Natural Language Processing (NLP) community in recent years \cite{ mai2023learning, wang2024wisdom}. Limited by various factors, early sentiment analysis could only be conducted through text, which is far from sufficient for the diversified information of the real world. The same sentence, accompanied by different facial expressions or tones, can convey entirely different emotions. Therefore, utilizing multimodal information to perceive human emotions can enable AI to make more accurate judgments.
	
	In recent years, the emergence of pre-trained Large Language Model (LLM) has introduced some new paradigms to the research community in NLP. Thanks to the powerful capabilities of LLM, they have demonstrated outstanding performance on many downstream tasks. Consequently, numerous researchers have begun to develop domain-specific LLM (such as for healthcare, weather forecasting, etc.) \cite{toma2023clinical, bi2023accurate}, by fine-tuning LLM with high-quality data to exhibit strong capabilities in relevant fields. This paradigm of LLM-for-specific-field is an excellent approach, but it demands high quality and quantity of data, making its implementation somewhat challenging.
	
	Some researchers in the field of sentiment analysis also leveraged this paradigm to tailor models specifically for MSA and ERC tasks. 
	Hu et al. \shortcite{hu2022unimse} introduced UniMSE, which for the first time unifies MSA and ERC as generative tasks. 
	UniMSE assigns emotional category labels to samples in MSA tasks and sentiment intensity labels to samples in ERC tasks by calculating textual similarity. 
	Subsequently, a model is trained based on the T5 (Raffel et al., 2020) architecture, exhibiting remarkable proficiency in both MSA and ERC tasks.
	Li et al. \shortcite{li2023unisa} developed UniSA, a comprehensive framework for sentiment analysis. Built upon the BART \cite{lewis2019bart} model and a series of pre-training tasks, UniSA excels in a wide range of multimodal and text-only sentiment analysis tasks, offering a more comprehensive performance compared to UniMSE.
	Although the two aforementioned works exhibit remarkable performance, they still exhibit certain limitations. 1) \textbf{Significant computational overhead}. Large number of trainable parameters and multi-task training lead to considerable computational expenses, where UniSA's pre-training requires three days on eight NVIDIA RTX V100 32G GPUs. 
	2) \textbf{Losing the generalization capability inherent to the base models}. UniMSE and UniSA, trained respectively on T5  \cite{raffel2020exploring} and BART \cite{lewis2019bart}, are tailored for the sentiment analysis domain, restricting these models from performing tasks outside of sentiment analysis. 

	Currently, there is a trend towards developing plugins that freeze the backbone LLM and train them to perform non-textual tasks while retaining their inherent capabilities \cite{tsimpoukelli2021multimodal, alayrac2022flamingo, chen2023x, sun2023test, szot2023large}
	Given the expectation of continued growth in the parameter size of LLM, we propose developing a plugin with reduced training overhead. 
	This will enable LLM to maintain its intrinsic capabilities while effectively executing a range of multimodal sentiment analysis tasks with minimal computational resources. We believe this approach is more promising than training a specialized sentiment analysis LLM.
	%
	Our contributions are summarized as follows:
	
	\begin{itemize}
		\item We proposed \textbf{M}ultimodal \textbf{S}entiment Analysis and \textbf{E}motion Recognition \textbf{Adapter} (MSE-Adapter), a plug-and-play lightweight plugin. This plugin is trained with minimal overhead without altering the inherent capabilities of the LLM, enabling it to perform MSA or ERC task via autoregressive generated sentiment labels.
		\item In the MSE-Adapter, we introduce the Text-Guide-Mixer (TGM) and Multi-Scale Fusion (MSF) modules. The TGM establishes explicit connections between non-textual and textual modalities. This facilitates feature-level alignment between them and encourages the LLM to better understand content from non-textual modalities. The MSF module enables early fusion of non-textual features across multiple scales, further augmenting the LLM's efficiency in integrating information from diverse modalities.
		\item Extensive experiments were conducted on four publicly available Chinese and English datasets using consumer-grade GPUs with open-source LLMs (Qwen-1.8B, ChatGLM3-6B-base, and LLaMA2-7B) as the backbone. The results demonstrate the effectiveness of the MSE-Adapter.
	\end{itemize}

	\section{Related Work}
	
	\subsection{Pre-trained language model for MSA and ERC}
	Currently, in the MSA and ERC communities, a number of outstanding works have emerged, including contrastive learning-based methods \cite{yang2023confede, mai2023learning}, graph-based methods \cite{li2023ga2mif, hu2021mmgcn, lin-etal-2022-modeling}, transformer-based methods \cite{sun2023layer, zhang2023learning, yang2023code}, and methods based on Pre-trained Language Model (PLM). PLM-based methods typically use a designated PLM, such as BERT \cite{devlin2018bert} or T5 \cite{raffel2020exploring}, as their foundation. These methods convert non-textual modality features into tokens with equivalent dimensions to those of textual modalities, enabling training within the PLM framework.

	Rahman et al. \shortcite{rahman2020integrating} proposed the MAG, where completed word embeddings are fused with non-textual modality features to generate new embeddings. These embeddings are then fine-tuned within PLMs (such as BERT and XLNet \cite{yang2019xlnet}) to achieve notable performance improvements.
	Similarly, Guo et al. \shortcite{guo2022dynamically} proposed CHFN, which uses its designed Multimodal Interaction layer to integrate non-textual modal information into textual embedding at the word level, and then fine-tunes the integrated multimodal information by feeding it into BERT.  
	%
	%
	Additionally, Hasan et al. \shortcite{hasan2023textmi} presented TextMI, a method that converts audio and vision information into corresponding textual descriptions. This approach links these descriptions with the textual content, transforming multimodal information into purely textual information. By inputting this enhanced text into BERT, TextMI achieves competitive performance results.
	Hu et al. \shortcite{hu2022unimse} introduced UniMSE, an approach that uses the T5 \cite{raffel2020exploring} model as its foundation. UniMSE encodes text using the initial layers of T5's encoder and then trains the remaining layers using a combination of non-textual and textual features. This training strategy incorporates contrastive learning to enhance the model's representation learning capabilities. Benefiting from its multitask training paradigm, UniMSE shows capabilities in both MSA and ERC tasks, demonstrating exceptional performance.
	%
	Li et al. \shortcite{li2023unisa} developed UniSA, a comprehensive framework for sentiment analysis. UniSA uses PLMs (GPT2-medium \cite{radford2019language}, T5 and BART) as a foundation, standardizing the data formats of various types of sentiment analysis sub-tasks for input into PLMs. It leverages pre-training tasks and contrastive learning to pre-train the PLM, followed by fine-tuning on downstream task datasets. The UniSA$_{\text{BART}}$ achieve comprehensive results across multiple sentiment analysis sub-tasks. 
	%
	
	Compared to aforementioned works, our proposed approach, MSE-Adapter, is a lightweight plugin that requires fewer training parameters (approximately 2.6M to 2.8M trainable parameters for base models sized 6/7B). Notably, MSE-Adapter preserves the inherent generalization capability of the LLM without sacrificing efficiency. Therefore, when assigned MSA or ERC tasks, the user can invoke the relevant pre-trained MSE-Adapter plugin to carry out the designated task. This design enhances parameter efficiency while preserving the effectiveness and adaptability of LLM, offering a new and efficient solution for using LLM in MSA and ERC tasks.
		
	\begin{figure*}[ht]
		\centering
		\includegraphics[width=0.78\linewidth,trim=163 76 102 56,clip]{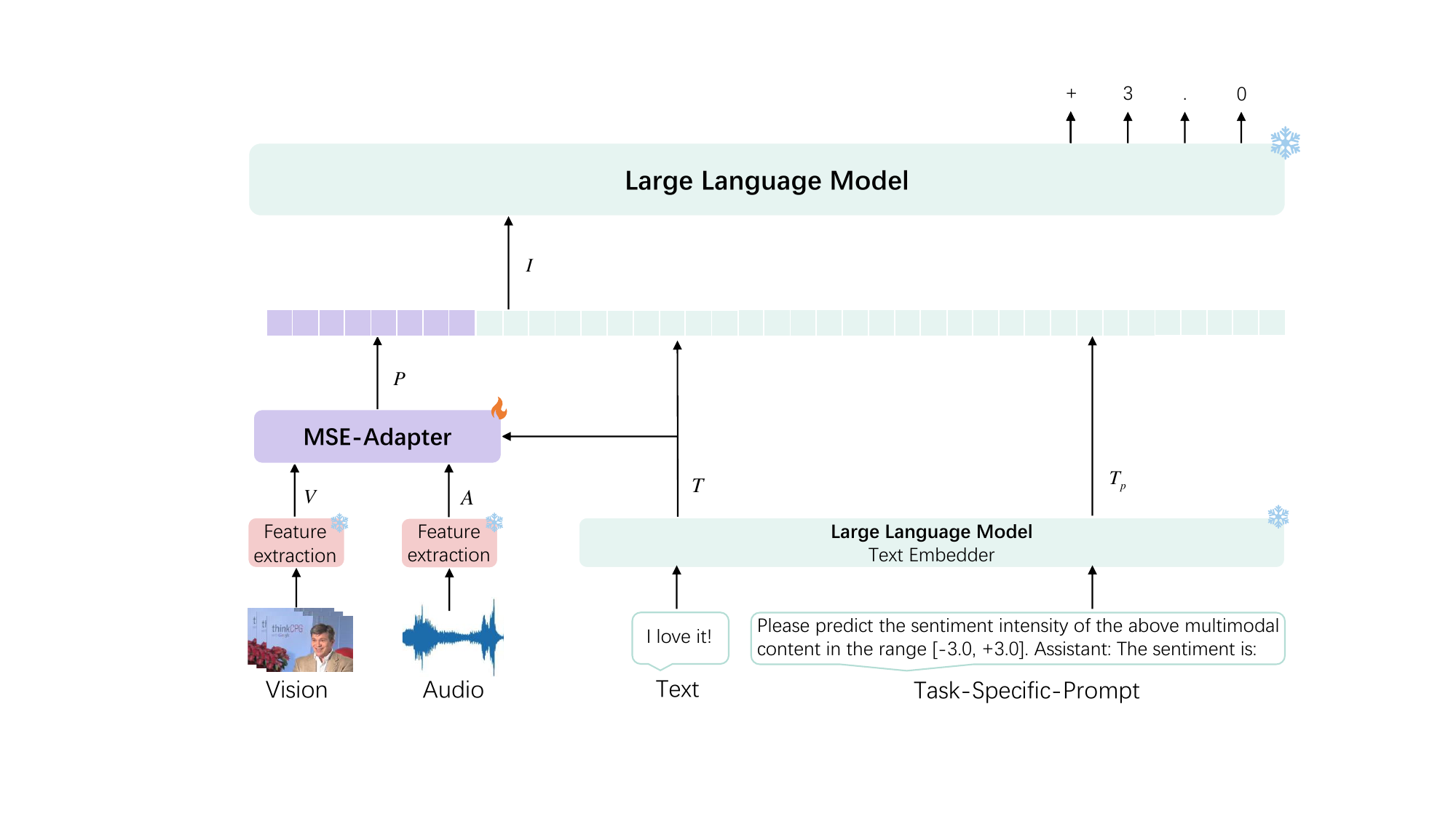}
		\caption{The comprehensive framework integrating MSE-Adapter with LLM.} 
		\label{figure 1}
	\end{figure*}

	\subsection{Adapters enabling LLM to perform non-plain text tasks}
	Adapters were usually utilized for efficiently fine-tuning large pre-trained models \cite{houlsby2019parameter, pfeiffer2020adapterhub, he2022sparseadapter, hu2023llm}. By freezing the main body of the pre-trained model and only training the Adapter, the essence is to use gradient backpropagation to let the Adapter generate pseudo tokens that can be recognized by the pre-trained model. This process is aimed at prompting the pre-trained models to further adapt to certain downstream tasks. Meanwhile, since the pre-trained model is frozen during the training phase, it retains its strong generalization capability, avoiding the issue of catastrophic forgetting \cite{liu2021p, liu2023gpt}. Inspired by these relevant works of Adapter, some researchers have argued that it is possible to convert information from non-textual modalities into information understandable by LLMs through an Adapter, enabling them to perform downstream tasks involving non-plain text modalities.
	Tsimpoukelli et al. \shortcite{tsimpoukelli2021multimodal} introduced Frozen, which utilizes a vision encoder to convert images into a series of tokens. These tokens are concatenated with a prompt and used to train a LLM for visual question answering (VQA) and captioning tasks, with gradient backpropagation guiding updates to the parameters of the vision encoder.
	%
	Similarly, Alayrac et al. \shortcite{alayrac2022flamingo} proposed Flamingo, which incorporates trainable cross-attention layers into a frozen LLM to fuse textual and vision modalities after embedding vision modality information using a pre-trained vision encoder. Flamingo exhibits remarkable performance across various video/visual-related tasks following training. 
	%
	Chen et al. \shortcite{chen2023x} presented X-LLM, a model that leverages X2L interfaces to convert vision, image, and audio modalities into ``foreign languages" that can be processed by the LLM. X-LLM demonstrates impressive performance after instruction-tuning on a high-quality multimodal instruction dataset.
	Sun et al. \shortcite{sun2023test} developed TEST, which utilizes contrastive learning to train an encoder for time series (TS) data, applies similarity constraints to align it with text, and fine-tunes the LLM with a soft-prompt approach to effectively process TS-related tasks.
	
	In this paper, we introduce a lightweight plugin named MSE-Adapter, which enable the LLM's to perform MSA or ERC task without affecting its inherent capabilities. Unlike previous works, we introduce a novel module named TGM in the MSE-Adapter. TGM facilitates feature-level alignment between non-textual and textual modalities, which aids the LLM to better understand content from non-textual modalities.

	\section{Method}
	
	\subsection{Overall architecture}
	Figure \ref{figure 1} presents the comprehensive framework integrating MSE-Adapter with LLM.
	Given a sample ${{x}_{i}}$, the first step is to convert each modality into a sequence embedding. For the textual modality, we use the Text Embedder within the LLM to tokenize \cite{kudo2018sentencepiece} the input from the text modality and the Task-Specific-Prompt, and then convert them into sequence embeddings. For audio and vision modalities, we employ pre-trained toolkits \cite{yu2021learning, liu2022make, hu2022unimse, sun2023layer} for feature extraction to transform them into feature sequence embeddings (See Appendix A for more details). After the encoding process, the sample's textual modality, Task-Specific-Prompt, vision, and audio modalities are represented as ${T\in {{\mathbb{R}}^{{{l}_{t}}\times {{d}_{t}}}}}$, ${{{T}_{p}}\in {{\mathbb{R}}^{{{l}_{t}}\times {{d}_{t}}}}}$, $V\in {{\mathbb{R}}^{{{l}_{\nu }}\times {{d}_{\nu }}}}$ and $A\in {{\mathbb{R}}^{{{l}_{a}}\times {{d}_{a}}}}$, respectively, where ${{l}_{m\in \{t,v,a\}}}$ denotes the sequence length of each modality embedding, and ${{d}_{m\in \{t,v,a\}}}$ represents the corresponding feature vector dimension. 
	
	Subsequently, $T$, $V$ and $A$ are fed into the MSE-Adapter for further processing and generating pseudo tokens $P$. Finally, we concatenate $P$, $T$, ${{T}_{p}}$ to obtain $I$ (i.e., $I=[P;T;{{T}_{p}}]$) and input it into the frozen LLM, which returns the logits ${{g}_{i}}$ corresponding to the input sample and the generated text ${{y}_{i}}$ for the entire sentence (including input and output tokens). This can be expressed as:
	\begin{equation}
		\{{y}_{i},{g}_{i}\}=LLM(I,\theta )
	\end{equation}
	where $\theta$ represents the parameters of the LLM. The LLM predicts the conditional probability $\rho({{\gamma }_{j}}|I,\theta )$ of each token ${{\gamma }_{j}}$ of the generated text ${{y}_{i}}$ until the end-of-sequence symbol $<$eos$>$. 
	For the logits ${{g}_{i}}\in {{\mathbb{R}}^{{{I}_{l}}\times {{V}_{s}}}}$, where ${{I}_{l}}$ and ${{V}_{s}}$ represent the length of the input $I$ and the size of the vocabulary used by the LLM, respectively.
	
	Following the original training methodology of the LLM, we utilize the next-token prediction loss to measure the output error of the model. Hence, the loss calculation for the model task ${{L}_{task}}$ is defined as follows:
	\begin{equation}
		{{L}_{task}}=\sum\limits_{k=1}^{N}{-log\rho ({{Y}_{k}}|I,\theta  )}
	\end{equation}
	where ${{Y}_{k}}$ represents the $k$-{th} token of the sentiment label corresponding to the sample ${{x}_{i}}$, $N$ denotes the number of sentiment label tokens. Based on the aforementioned loss, we optimize the parameters of the MSE-Adapter through gradient backpropagation, thereby enhancing the MSE-Adapter's adaptation to the LLM. Upon training completion, the LLM can utilize the MSE-Adapter to receive multimodal inputs and generate autoregressive sentiment labels, similar to text generation.
	
	\begin{figure}[ht]
		\centering
		\includegraphics[width=0.673\linewidth,trim=0 15 700 0,clip]{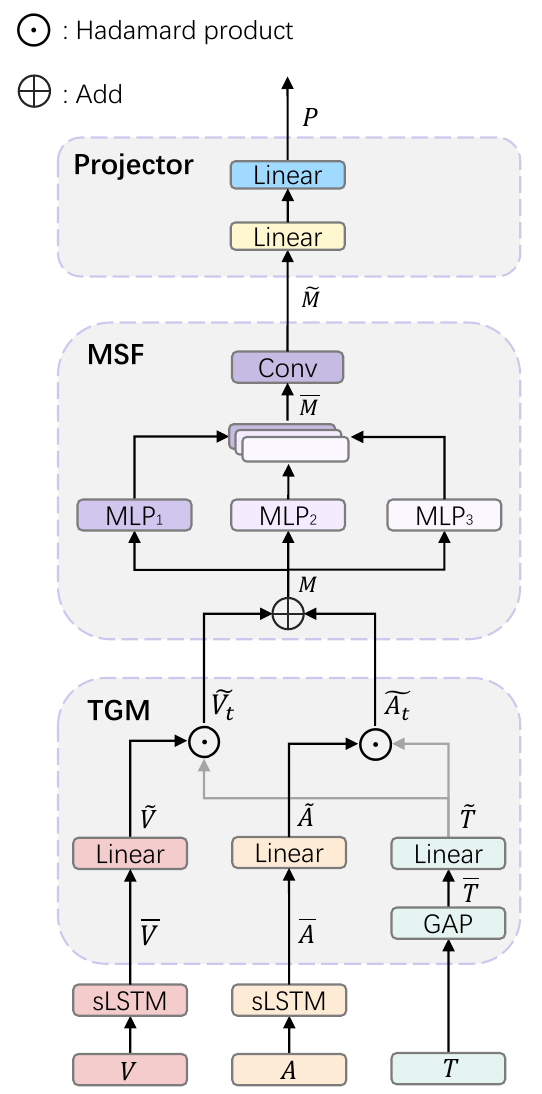}
		\caption{The architecture of MSE-Adapter.} 
		\label{figure 2}
	\end{figure}
	
	\subsection{MSE-Adapter}
	In this section, we introduce the proposed MSE-Adapter, whose structure is depicted in Figure \ref{figure 2}. The MSE-Adapter consists of two separate single directional Long Short-Term Memory (sLSTM) modules, the TGM module, the MSF module, and a Projector module. 
	The sLSTM initially performs temporal modeling separately for $V$ and $A$, and then it captures their end-state hidden representations to derive $\overline{V}\in {{\mathbb{R}}^{{{h}_{v}}\times 1}}$ and $\overline{A}\in {{\mathbb{R}}^{{{h}_{a}}\times 1}}$.
	These outputs are then fed into the TGM to obtain $\widetilde{{{V}_{t}}}\in {{\mathbb{R}}^{h\times 1}}$ and $\widetilde{{{A}_{t}}}\in {{\mathbb{R}}^{h\times 1}}$. Subsequently, these are input into the MSF for fusion, resulting in $\widetilde{M}\in {{\mathbb{R}}^{h\times 1}}$. Finally, a Projector composed of two linear layers expands this into pseudo tokens $P$.
	
	\subsubsection{\textbf{Text-Guide-Mixer}}
	Feature alignment across different modalities has always been a significant issue in multimodal tasks. A key to modal feature alignment is establishing connections between modalities. In the research on recommendation systems and search engines, researchers have applied the Hadamard product to achieve feature crossing, thus creating explicit connections between features at the vector level \cite{lian2018xdeepfm,wang2021dcn}. Inspired by these works, we propose the TGM module. TGM establishes an explicit connection by computing the Hadamard product between the feature vectors of the text modality and those of the non-text modalities. 
	This strategy not only preserves the original individual features of the non-text modalities but also encourages non-text feature vectors to align with text modality feature vectors, narrowing the gap between non-text modality features and text features, enabling the MSE-Adapter to generate pseudo tokens that are more easily understood by LLM. 
	
	The implementation is as follows: we first do global average pooling (GAP) on the textual modal inputs $T$ to obtain $\overline{T}\in {{\mathbb{R}}^{1\times {{d}_{t}}}}$, and then we use a linear layer to project $\overline{T}$, $\overline{V}$ and $\overline{A}$ to the same dimension:
	\begin{equation}
		\begin{aligned}
			\widetilde{T} &= W_{t}\overline{T}^{T} \\ 
			\widetilde{V} &= W_{v}\overline{V} \\ 
			\widetilde{A} &= W_{a}\overline{A}
		\end{aligned}
	\end{equation}
	where ${{W}_{t}}\in {{\mathbb{R}}^{h\times {{d}_{t}}}}$, ${{W}_{v}}\in {{\mathbb{R}}^{h\times {{h}_{v}}}}$, and ${{W}_{a}}\in {{\mathbb{R}}^{h\times {{h}_{a}}}}$.
	
	Subsequently, we perform the Hadamard product of $\widetilde{V}$ and $\widetilde{A}$ with $\widetilde{T}$, respectively:
	\begin{equation}
		\begin{aligned}
			& \widetilde{{{V}_{t}}}=\widetilde{V}\odot \widetilde{T} \\ 
			& \widetilde{{{A}_{t}}}=\widetilde{A}\odot \widetilde{T} \\ 
		\end{aligned}
	\end{equation}
	where \(\odot\) represents the Hadamard product, that is, the element-wise multiplication of matrices. Through this process, we obtain the new non-textual modality representations $\widetilde{{{V}_{t}}}$ and $\widetilde{{{A}_{t}}}$.
	
	\subsubsection{\textbf{Multi-Scale-Fusion}}
	Entrusting the task of complete multimodal fusion to a frozen LLM presents certain challenges. In response to this challenge, we adopt an early fusion approach for the non-textual modalities prior to their input into the LLM. Specifically, we introduc a module named MSF, dedicated to performing low-level fusion of the non-textual modalities. Subsequently, the high-level fusion between the textual and non-textual modalities is deferred to the LLM. 
	This layered fusion approach enables the LLM to capture more refined and detailed multimodal fusion, thereby boosting model performance.
	
	The implementation of the MSF module is as follows: Firstly, we sum the outputs $\widetilde{{{V}_{t}}}$ and $\widetilde{{{A}_{t}}}$ derived from the TGM to obtain $M$. Subsequently, we utilize three Multi-Layer Perceptrons (MLPs) with diverse hidden layer dimensions to conduct feature fusion at multiple scales on $M$. Specifically, for the \(i\)th $(i\in \{1,2,3\})$ MLP, we obtain \(m_i\):
	
	
	\begin{equation}
		{{m}_{i}}=W_{2}^{i}\sigma (W_{1}^{i}M)
	\end{equation}
	where $W_{1}^{i}\in {{\mathbb{R}}^{h/k\times h}},W_{2}^{i}\in {{\mathbb{R}}^{h\times h/k}},k\in \{8,16,32\}$, $\sigma $ represents the GELU activation function.
	Subsequently, we stack the fusion results from the three different scales to obtain  $\overline{M}=[{{m}_{1}},{{m}_{2}},{{m}_{3}}]\in {{\mathbb{R}}^{h\times 3}}$. To further integrate information from various scales, we use a 1×1 convolution (Conv) to compress this information, resulting in $\widetilde{M}\in {{\mathbb{R}}^{h\times 1}}$.
	
	\subsubsection{\textbf{Projector}}
	After extracting the fused information from non-textual modalities, the size of $\widetilde{M}$ is adjusted to meet the input requirements of the LLM using a linear layer. Subsequently, another linear layer is utilized to increase the number of pseudo tokens:
	\begin{equation}
		P={{W}_{4}}({{W}_{3}}\widetilde{M})^T
	\end{equation}
	where ${{W}_{3}}\in {{\mathbb{R}}^{{{d}_{t}}\times h}},{{W}_{4}}\in {{\mathbb{R}}^{n\times 1}}$ ($n$ is our predetermined number of pseudo tokens). 
	The resulting pseudo tokens, denoted by $P$, can then be utilized as input into the LLM.
	
	\section{Experiment}
	\subsection{Datasets}
	To more comprehensively evaluate the performance of the MSE-Adapter, we conducted MSA and ERC experiments on two popular English datasets and two popular Chinese datasets, namely MOSEI \cite{zadeh2018multimodal}, MELD \cite{poria2018meld}, SIMS-V2 \cite{liu2022make}, and CHERMA \cite{sun2023layer}. The details of the dataset are shown in the appendix B.

	\subsection{Experimental Settings}
	In this subsection, we briefly introduce the detailed setup of our experiments. Our investigation utilizes the Qwen-1.8B \cite{bai2023qwen}, ChatGLM3-6B-base \cite{du2022glm}, and LLaMA2-7B \cite{touvron2023llama} models as the backbone. For ease of presentation, we add the prefix `MSE' to the backbone to indicate the integration of the MSE-Adapter (e.g., MSE-Qwen-1.8B). The trainable parameters of MSE-Adapter in various LLMs are presented in Table \ref{table1}. For a fair comparison, we selected [1111, 2222, 3333, 4444, 5555] as random seeds for the experiments and reported the average results achieved across these five random seeds. All experiments were conducted on a single NVIDIA RTX 4090 GPU. The optimizer used was AdamW with a warmup learning rate strategy. The rest of the settings can be found in Appendix C.  
	
	\begin{table}[h]
		\centering
		\small 
		\setlength{\tabcolsep}{2.5pt} 
		\renewcommand{\arraystretch}{1.1} 
		\begin{tabular}{ccccc}
			\hline\hline
			\multirow{2}{*}{Model} & \multicolumn{4}{c}{Trainable parameters}  \\ 
			\cline{2-5}
			& MOSEI & SIMS-V2 & MELD  & CHERMA          \\ 
			\hline
			\textbf{MSE-Qwen-1.8B}     & 1.35M & 1.40M   & 1.35M & 1.56M           \\
			\textbf{MSE-LLaMA2-7B}     & 2.63M & 2.68M   & 2.63M & 2.84M           \\
			\textbf{MSE-ChatGLM3-6B}   & 2.63M & 2.68M   & 2.63M & 2.84M           \\
			\hline\hline   
		\end{tabular}
		\normalsize
		\caption{Trainable parameters of LLMs integrating MSE-Adapter.}
		\label{table1}
	\end{table}

	To standardize labels for the MSA task, for models where the tokenizer does not automatically generate tokens to distinguish between positive and negative labels, we manually add a `+' sign to labels greater than or equal to 0. 
	Furthermore, to streamline the answer generation process for the ERC task, we translated the emotion labels into distinct numerical values and incorporated these details into the prompts, as illustrated in Appendix C.

	\subsection{Baselines}
	
	We compared the performance of ``MSE-Qwen-1.8B", ``MSE-LLaMA2-7B" and ``MSE-ChatGLM3-6B" to that of previous state-of-the-art models: TFN \cite{zadeh2017tensor}, LMF \cite{liu2018efficient}, MISA \cite{hazarika2020misa}, MAG-BERT \cite{rahman2020integrating}, Self-MM \cite{yu2021learning}, MMIM \cite{han2021improving}, CHFN \cite{guo2022dynamically}, UniMSE \cite{hu2022unimse}, UniSA$_{\text{BART}}$, UniSA$_{\text{T5}}$, UniSA$_{\text{GPT2}}$ \cite{li2023unisa}, AV-MC \cite{liu2022make}, MMGCN \cite{hu2021mmgcn}, MM-DFN \cite{hu2022mm}, EmoCaps \cite{li2022emocaps}, GA2MIF \cite{li2023ga2mif}, EFT, LFT, MulT \cite{tsai2019multimodal}, PMR \cite{lv2021progressive}, and LFMIM \cite{sun2023layer}. The details of these baseline models are given in the Appendix D.
	
	\subsection{Metrics}
	In this study, due to the differing labels across datasets, we report our experimental results using a variety of metrics tailored to each dataset. For MOSEI, we report the mean absolute error (MAE), Pearson correlation (Corr), seven-category accuracy (Acc-7), binary accuracy (Acc-2), and F1 score as evaluation metrics (where Acc-2 and F1 are calculated based on a non-negative/negative standard). For SIMS-V2, we report MAE, Corr, Acc2\_weak, Acc-2, and F1 (where Acc-2 and F1 are calculated based on a non-positive/positive standard, and Acc2\_weak is used to further validate the model's performance on weakly emotional instances within the [-0.4, 0.4] range). For MELD and CHERMA, we report seven-category accuracy (Acc) and weighted F1 (WF1).

	\subsection{Results}
	Tables \ref{table2} and \ref{table3}  present the results of the performance comparison of multiple methods on the MSA and ERC tasks, respectively. It is noteworthy that, overall, all three LLMs incorporating the MSE-adapter exhibited outstanding performance.
	
	\begin{table*}[!h]
		\centering
		\renewcommand{\arraystretch}{1.1} 
		\begin{tabular}{cccccc|ccccc} 
			\hline\hline
			\multirow{2}{*}{Model} & \multicolumn{5}{c|}{MOSEI}                                                         & \multicolumn{5}{c}{SIMS-V2}                                                         \\ 
			\cline{2-11}
			& Acc-2          & F1             & Acc-7          & MAE            & Corr           & Acc-2          & F1             & Acc2\_weak     & MAE            & Corr            \\ 
			\hline
			TFN*                   & 78.50           & 78.96          & 51.60           & 0.573          & 0.714          & 76.51          & 76.31          & 66.27          & 0.323          & 0.667           \\
			LMF*                   & 80.54          & 80.94          & 51.59          & 0.576          & 0.717          & 77.05          & 77.02          & 69.34          & 0.343          & 0.638           \\
			MulT*                  & 81.15          & 81.56          & 52.84          & 0.559          & 0.733          & 79.50           & 79.59          & 69.61          & 0.317          & 0.703           \\
			MAG-BERT*              & 82.51          & 82.77          & 50.41          & 0.583          & 0.741          & 79.79          & 79.78          & 71.87          & 0.334          & 0.691           \\
			MISA                   & 83.60           & 83.80           & 52.20           & 0.555          & 0.756          & 80.53          & 80.63          & 70.50           & 0.314          & 0.725           \\
			Self-MM*               & 82.81          & 82.53          & 53.46          & 0.530           & 0.765          & 79.01          & 78.89          & 71.87          & 0.335          & 0.640            \\
			MMIM                   & 82.24          & 82.66          & 54.24          & 0.526          & 0.772          & 80.95          & 80.97          & 72.28          & 0.316          & 0.707           \\
			AV-MC                  & -              & -              & -              & -              & -              & 82.50           & 82.55          & 74.54          & 0.297          & \textbf{0.732}  \\
			CHFN                 & 83.70          & 83.90          & 54.30          & 0.525          & 0.778          & -              & -              & -              & -              & -               \\
			UniMSE                 & 85.86          & 85.79          & 54.39          & 0.523          & 0.773          & -              & -              & -              & -              & -               \\
			UniSA$_{\text{GPT2}}$                  & 71.02          & -              & 41.36          & 0.838          & -          & -              & -              & -              & -              & -               \\ 
			UniSA$_{\text{T5}}$                  & 84.22          & -              & 52.50          & 0.546          & -          & -              & -              & -              & -              & -               \\ 
			UniSA$_{\text{BART}}$                  & 84.93          & -              & 50.03          & 0.587          & -          & -              & -              & -              & -              & -               \\ 
			\hline
			\textbf{MSE-Qwen-1.8B}     & 84.12          & 83.45          & 52.02          & 0.558          & 0.725          & 80.44          & 80.24          & 73.09          & 0.311          & 0.678           \\
			\textbf{MSE-LLaMA2-7B}     & 86.74          & 86.51          & \textbf{55.57} & \textbf{0.501} & \textbf{0.787} & 75.53          & 75.44          & 68.61          & 0.382          & 0.553           \\
			\textbf{MSE-ChatGLM3-6B}   & \textbf{86.91} & \textbf{86.77} & 54.56          & 0.515          & 0.783          & \textbf{83.77} & \textbf{83.76} & \textbf{75.24} & \textbf{0.296} & 0.720            \\
			\hline\hline
		\end{tabular}
		\caption{Experimental results of the MSA task on the MOSEI and SIMS-V2 datasets: 1) Results for models marked with * on MOSEI are sourced from the official repository\protect\footnotemark, while results for other models are extracted from relevant published papers; 2) All models' results on SIMS-V2 are cited from literature \cite{liu2022make}.}
		\label{table2}
		
	\end{table*}
	\footnotetext{https://github.com/thuiar/MMSA/blob/master/results/result-stat.md}
	
	\begin{table}[htbp]
		\centering
		\renewcommand{\arraystretch}{1.1} 
		\begin{tabular}{ccc|cc} 
			\hline\hline
			\multirow{2}{*}{Model} & \multicolumn{2}{c|}{MELD}       & \multicolumn{2}{c}{CHERMA}      \\ 
			\cline{2-5}
			& Acc            & WF1            & Acc           & WF1             \\ 
			\hline
			TFN*                   & 60.77          & 57.74          & -             & 68.37           \\
			LMF*                   & 61.15          & 58.30           & -             & 68.23           \\
			EFT                    & -              & -              & -             & 68.72           \\
			LFT                    & -              & -              & -             & 69.05           \\
			MulT                   & -              & -              & -             & 69.24           \\
			PMR                    & -              & -              & -             & 69.53           \\
			LFMIM                  & -              & -              & -             & 70.54           \\
			MMGCN*                 & 60.42          & 58.31          & -             & -               \\
			MM-DFN*                & 62.49          & 59.46          & -             & -               \\
			EmoCaps                & -              & 64.00             & -             & -               \\
			GA2MIF                 & 61.65          & 58.94          & -             & -               \\
			UniMSE                 & 65.09          & \textbf{65.51} & -             & -               \\
			UniSA$_{\text{GPT2}}$                  & 48.12          & 31.26          & -             & -               \\ 
			UniSA$_{\text{T5}}$                  & 64.52          & 62.17          & -             & -               \\ 
			UniSA$_{\text{BART}}$                  & 62.34          & 62.22          & -             & -               \\ 
			\hline
			\textbf{MSE-Qwen-1.8B}     & 62.18          & 59.87          & 70.38         & 70.21           \\
			\textbf{MSE-LLaMA2-7B}     & 65.14          & 63.66          & 71.58         & 71.41           \\
			\textbf{MSE-ChatGLM3-6B}   & \textbf{66.23} & 65.13          & \textbf{72.90} & \textbf{72.73}  \\
			\hline\hline
		\end{tabular}
		\caption{Experimental results of the ERC task on the MELD and CHERMA datasets: 1) Results for models marked with * on MELD are cited from the 
			the literature \cite{hu2022mm}, while results for other models are extracted from relevant published papers; 2) All models' results on CHERMA are cited from literature \cite{sun2023layer}.}
		\label{table3}
	\end{table}

	It is evident that MSE-ChatGLM3-6B, achieved the most comprehensive performance, outperforming baseline models on most metrics. Notably, it showed a great performance improvement on CHERMA, the dataset with the largest volume of data, where its WF1 was 2.19\% higher than the best baseline LFMIM. Although MSE-LLaMA2-7B has a larger parameter size than MSE-ChatGLM3-6B, its overall performance was not as good, especially on the Chinese datasets SIMS-V2. This might be due to MSE-LLaMA2-7B's inherent limitations in processing Chinese text. Since the LLM is frozen, the performance of the model is highly dependent on the inherent capabilities of the LLM. 
	
	Interestingly, MSE-LLaMA2-7B still performed better than the baseline on CHERMA (even though it is not particularly skilled in Chinese), somewhat confirming that LLMs are essentially pattern machines \cite{generalpatternmachines2023}, capable of learning mapping relationships from large data volumes, even for tasks they are less proficient in. This was observed similarly on MSE-Qwen-1.8B. Although MSE-Qwen-1.8B's performance was lower than the other two backbones, it requires the least trainable parameters. Since MSE-Qwen-1.8B is deployable on mobile devices, such a small-parameter plugin can offer users a more efficient interaction experience.

	\section{Discussion}
	\subsection{Ablation Study}
	To evaluate the contribution of each module within the MSE-Adapter, we conducted ablation experiments on the English dataset MELD and the Chinese dataset SIMS-V2. For MELD, we reported Acc and WF1 score, and for SIMS-V2, we reported Acc-2 and F1 score. Given that MSE-ChatGLM3-6B achieved the most comprehensive performance in our experiments, all of our ablation experiments were based on MSE-ChatGLM3-6B and reported the average results achieved across the five random seeds  (same random seeds as experimental). Table \ref{table4} shows the results of the ablation experiments.
	
	\begin{table}
		\centering
		\renewcommand{\arraystretch}{1.1} 
		\begin{tabular}{ccc|cc} 
			\hline\hline
			\multirow{2}{*}{} & \multicolumn{2}{c|}{MELD}        & \multicolumn{2}{c}{SIMS-V2}       \\ 
			\cline{2-5}
			& Acc            & WF1            & Acc-2           & F1              \\ 
			\hline
			w/o
			$A$           & 66.24          & 65.09          & 83.46           & 83.42~          \\
			w/o
			$V$           & 66.13          & 65.00          & ~78.84~         & 78.68~          \\
			w/o
			$T$           & 46.44          & 36.53          & 72.78           & 72.21           \\
			w/o
			$A, V$        & 57.11          & 53.92          & ~76.54          & 75.98~          \\ 
			\hline
			w/o
			TGM         & 65.07          & 63.85          & ~82.98          & 82.97~          \\
			w/o
			MSF         & 62.72          & 61.75          & ~81.37          & 81.32~          \\ 
			\hline
			MSE-Adapter       & \textbf{66.23} & \textbf{65.13} & \textbf{~83.77} & \textbf{83.76}  \\
			\hline\hline
		\end{tabular}
		\caption{The ablation experiments results on MELD and SIMS-V2, where $T$, $V$, and $A$ represent textual, vision, and audio modalities, respectively.  And the ‘w/o’ means remove a modality or module. Furthermore, in the case of ``w/o $T$", the TGM module becomes ineffective. $\overline{V}$ and $\overline{A}$ are mapped to the same dimension through two independent linear layers and then directly summed before being fed into the MSF module. In the case of ``w/o $A$, $V$", $\overline{V}$ and $\overline{A}$ are randomly initialised into the corresponding dimensions into the TGM and the model is trained anyway.}  
		\label{table4}
	\end{table}

	\subsubsection{\textbf{Effectiveness of TGM and MSF}}

	Initially, we evaluated the TGM and MSF module within the MSE-adapter. The results shown in Table \ref{table4} indicated that removing TGM decreased model performance, highlighting the importance of establishing cross-modal connections between non-textual and textual features for modality alignment and model performance. Similarly, omitting MSF also resulted in a decline in performance, with a more significant impact than that of TGM. This observation further illustrates that the multimodal fusion capacity of frozen LLM is limited. Therefore, to enhance the fusion of the three modalities, early fusion of non-textual modalities prior to LLM input is necessary.
	

	\subsubsection{The Impact of Absent Modalities}
	Additionally, we removed one or several modalities from the multimodal signals to verify their impact on model performance. The ablation results are presented in Table \ref{table4}. From the results, we discovered that eliminating either the vision or audio modalities, or both, led to a decrease in performance, indicating the necessity of non-textual modalities (i.e., vision and audio) for solving MSA and ERC tasks. Notably, the removal of the textual modality had the most severe impact on performance, underscoring the critical role of language in identifying emotions in the real world.
	
	\subsection{Further Discussion with Models Adapted for BERT}
	
	In previous work, MAG \cite{rahman2020integrating} and CHFN \cite{guo2022dynamically} conducted similar research, primarily utilizing their designed adaptation model to fine-tune BERT with a classification header for MSA tasks. We previously reported their performance metrics. For a fair comparison, we reproduced their adaptation model based on the original paper and adapted it to ChatGLM3-6B within our framework. We conducted experiments on the MOSEI, MELD, SIMS-V2, and CHERMA dataset, maintaining consistent experimental settings with the aforementioned chapters. It should be noted that the hyperparameters are only reported in the original paper for the MOSEI dataset in the MSA task. Consequently, in our reproduction, we maintain consistency with the original paper by keeping the hyperparameters on MOSEI only. The hyperparameters on the rest of the datasets are set with reference to MOSEI. Due to space limitations, we only show partial results for MOSEI and MELD in table \ref{table5}, and complete results are detailed in Appendix E. 
	
	
	The experimental results demonstrate that the MSE-Adapter outperforms both MAG and CHFN when using the same backbone network.  
	Additionally, while replacing BERT with LLM in MAG and CHFN can improve performance, the improvements are limited.
	Despite integrating more comprehensive multimodal information, MAG and CHFN introduce a degree of information redundancy. When the backbone network is trained, it benefits from this rich information, which enhances its performance. However, when the backbone network is frozen, this redundancy complicates its ability to effectively interpret multimodal content. 
	%
	In contrast, the MSE-Adapter is more efficient in capturing key information and extracting key features used to complete sentiment analysis, which reduces the difficulty of the backbone's comprehension. In conclusion, MSE-Adapter not only effectively reduces the number of trainable parameters, but also enhances the performance of the backbone on MSA or ERC tasks.
	
	\begin{table}
		\centering
		\small 
		\setlength{\tabcolsep}{3.5pt} 
		\renewcommand{\arraystretch}{1.1} 
		\begin{tabular}{ccc|cc|c} 
			\hline\hline
			\multirow{2}{*}{Model} & \multicolumn{2}{c|}{MOSEI}      & \multicolumn{2}{c|}{MELD}       & \multirow{2}{*}{\faFire Paras}  \\ 
			\cline{2-5}
			& Acc-2          & F1             & Acc            & WF1            &                         \\ 
			\hline
			MAG-ChatGLM3-6B           & 85.10           & 84.73          & 60.38          & 59.81          &34.47M                \\
			CHFN-ChatGLM3-6B          & 85.58          & 85.26          & 63.03          & 62.08          &50.79M                \\ 
			\hline
			MSE-ChatGLM3-6B          & \textbf{86.91} & \textbf{86.77} & \textbf{66.23} & \textbf{65.13} & \textbf{2.63M}          \\
			\hline\hline
		\end{tabular}
		\normalsize
		\caption{Experimental results on the MOSEI and MELD datasets. The ``\faFire Paras" represents the trainable parameter of the model. }
		\label{table5}
	\end{table}

	\section{Conclusion}
	This paper presents the MSE-Adapter, a lightweight plug-and-play plugin that empowers an LLM to handle MSA or ERC task without compromising its inherent capabilities. The MSE-Adapter includes a module called TGM, which is designed to facilitate the alignment of non-textual modalities with textual ones. By employing the Hadamard product, TGM establishes explicit connections between non-textual and textual modalities at the feature level, thereby enhancing the LLM's comprehension of content from non-textual sources.  MSF is another module of MSE-Adapter that employs MLPs of varying scales for early fusion of non-textual modalities, followed by further integration of this information using convolutional layers. This further enhances the efficiency of LLM in fusing information from diverse modalities.
	Deployable on consumer-grade GPUs and utilizing open-source LLMs (Qwen-1.8B, ChatGLM3-6B-base, and LLaMA2-7B) as the backbone, we conducted extensive experiments across four public English and Chinese datasets. The competitive results demonstrate the efficacy of MSE-Adapter. It not only optimizes parameter efficiency but also maintains the LLM's efficiency and flexibility, providing a
	new solution and baseline for the application of LLM in MSA and ERC.
\section*{Acknowledgements}

This work was supported by the National Natural Science Foundation of China (Nos. 62450100 and U20A20224).

\bibliography{reference}

\appendix
\section{APPENDIX A}
\label{appendixA}
\subsection{Feature Extraction}
To ensure a fair comparison, our feature extraction for non-textual modalities follows the mainstream extraction approach of previous work in the community. However, due to differences in the organisation and year of the dataset release, feature extraction was done differently, and we will describe them separately.

\textbf{MOSEI}: For MOSEI, we followed the settings of the extracted data from the previous work \cite{yu2021learning} and used the unaligned version of its data. For visual features, OpenFace was used directly to locate faces and then further extracted to obtain feature sequences of 35-dimensions, and for audio features, COVAREP was used to extract feature sequences of 74-dimensions.

\textbf{SIMS-V2}: For SIMS-V2, we follow the settings in the dataset publisher and use the unaligned version of the data. For visual features, TalkNet \cite{tao2021someone} was used for face localisation, and then OpenFace was used for facial feature extraction, yielding a feature sequence of 177-dimensions. For audio features, they were extracted by the OpenSMILE \cite{eyben2010opensmile} backend at a sampling rate of 16000 Hz to obtain a feature sequence of 25-dimensions. 

\textbf{MELD}: For MELD, we followed the settings for extracting data from UNIMSE \cite{hu2022unimse} and UNISA \cite{li2023unisa}. For visual features, Effecientnet \cite{tan2019efficientnet} with supervised pre-training on the VGGface\footnote{\url{https://www.robots.ox.ac.uk/ vgg/software/vgg_face/}} and AFEW datasets was used to obtain video feature sequence of 64-dimensions. For audio features, the raw acoustic input was processed into digital sequence vectors via librosa\footnote{\url{https://github.com/librosa/librosa.}} to extract Mel spectrograms as audio feature feature sequence of 64-dimensions.

\textbf{CHERMA}: For CHERMA, we followed the settings in the dataset publisher \cite{sun2023layer}. For visual features, the video is first processed with MTCNN \cite{zhang2016joint} to obtain aligned faces, and then each frame is fed to the pre-trained Resnet 18 (trained with RAF-DB \cite{li2017reliable}), which outputs feature sequence of 512-dimensions. For audio features, the extracted frame-level features are fed into the pre-trained wav2vec \cite{zhang2022wenetspeech}, which generates a feature sequence of 768-dimensions.

\section{APPENDIX B}
\label{appendix b}
\subsection{The Dataset details}
In this section we give a brief description of the different datasets. Table \ref{table1} presents the data statistics and task types of these four datasets. 

\textbf{MOSEI}: The MOSEI dataset is an expanded version of MOSI \cite{zadeh2016mosi}, comprising 22,856 annotated video segments across more than 250 different topics. Similar to MOSI, each segment is scored for sentiment on a scale ranging from -3 (strongly negative) to +3 (strongly positive).

\textbf{SIMS-V2}: SIMS-V2 is an upgraded version of SIMS \cite{yu2020ch}, a Chinese multimodal dataset featuring 4,403 detailed video segments. Each sample is assigned a multimodal label along with three unimodal labels, with sentiment scores ranging from -1 (strongly negative) to +1 (strongly positive).

\textbf{MELD}: MELD is a multi-party dataset that includes over 1,400 dialogues and 13,708 utterances extracted from the TV series "Friends." The dataset categorizes each utterance into one of seven emotional states: neutral, surprise, fear, sadness, joy, disgust, and anger.

\textbf{CHERMA}: CHERMA is a Chinese emotion recognition dataset, composed of content from 148 TV dramas, 7 variety shows, and 2 movies, resulting in 28,717 extracted segments. Similar to MELD, this dataset categorizes each utterance into one of seven emotional states: neutrality, surprise, fear, sadness, happiness, disgust, and anger.

\begin{table}[h]
	\centering
	\small
	\setlength{\tabcolsep}{4pt} 
	\renewcommand{\arraystretch}{1.1} 
	\begin{tabular}{ccccccc} 
		\hline\hline
		\textbf{Dataset} & \textbf{Train} & \textbf{Valid} & \textbf{Test} & \textbf{Total} & \textbf{Task} & \textbf{Language}  \\ 
		\hline
		\textbf{MOSEI}   & 16326          & 1871           & 4659          & 22856          & MSA           & English            \\
		\textbf{SIMS-V2} & 2722           & 647            & 1034          & 4403           & MSA           & Chinese            \\
		\textbf{MELD}    & 9989           & 1109           & 2610          & 13708          & ERC           & English            \\
		\textbf{CHERMA}  & 17230          & 5743           & 5744          & 28717          & ERC           & Chinese            \\
		\hline\hline
	\end{tabular}
	\normalsize
	\caption{The statistics of MOSEI, SIMS-V2, MELD and CHERMA.}
	\label{table1}
\end{table}

\section{APPENDIX C}
\label{appendix c}
\subsection{Experimental hyperparameters}
For Qwen-1.8B\footnote{\url{https://huggingface.co/Qwen/Qwen-1_8B}}, ChatGLM3-6B-Base\footnote{\url{https://huggingface.co/THUDM/chatglm3-6b-base}}, and LLaMA2-7B\footnote{\url{https://huggingface.co/meta-llama/Llama-2-7b-hf}}, we used open-source pre-trained weights available on HuggingFace. The other hyperparameters are shown in Table \ref{table2}, \ref{table3} and \ref{table4}. The prompt on different datasets is shown in Figure \ref{figure 2}.

\begin{table}[h]
	\centering
	\small 
	\setlength{\tabcolsep}{2.5pt} 
	\renewcommand{\arraystretch}{1.1} 
	\arrayrulecolor{black}
	\begin{tabular}{ccccc} 
		\arrayrulecolor{black}\hline
		\textbf{MSE-Qwen-1.8B}   & MOSEI & SIMS-V2 & MELD  & CHERMA  \\ 
		\hline
		A\_LSTM\_hidden dim  & 64    & 64      & 32    & 32      \\
		V\_LSTM\_hidden dim  & 32    & 64      & 16    & 16      \\
		Learing\_Rate        & 5e-3  & 5e-4    & 5e-4  & 5e-3    \\ 
		Pseudo tokens num    & 4     & 4       & 2     & 4       \\
		\hline
	\end{tabular}
	\normalsize
	\caption{Hyperparameter settings for MSE-Qwen-1.8B.}
	\label{table2}
	\arrayrulecolor{black}
\end{table}

\begin{table}[h]
	\centering
	\small 
	\setlength{\tabcolsep}{2.5pt} 
	\renewcommand{\arraystretch}{1.1} 
	\arrayrulecolor{black}
	\begin{tabular}{ccccc} 
		\arrayrulecolor{black}\hline
		\textbf{MSE-ChatGLM3-6B} & MOSEI & SIMS-V2 & MELD  & CHERMA  \\ 
		\hline
		A\_LSTM\_hidden dim       & 64    & 64      & 64    & 32      \\
		V\_LSTM\_hidden dim       & 32    & 64      & 32    & 16      \\
		Learing\_Rate             & 5e-5  & 5e-5    & 5e-5  & 5e-5    \\ 
		Pseudo tokens num         & 4     & 4       & 4     & 4       \\
		\hline
	\end{tabular}
	\normalsize
	\caption{Hyperparameter settings for MSE-ChatGLM3-6B.}
	\label{table3}
	\arrayrulecolor{black}
\end{table}

\begin{table}[h]
	\centering
	\small 
	\setlength{\tabcolsep}{2.5pt} 
	\renewcommand{\arraystretch}{1.1} 
	\arrayrulecolor{black}
	\begin{tabular}{ccccc} 
		\arrayrulecolor{black}\hline
		\textbf{MSE-LLaMA-7B}   & MOSEI & SIMS-V2 & MELD  & CHERMA  \\ 
		\hline
		A\_LSTM\_hidden dim  & 64    & 64      & 64    & 32      \\
		V\_LSTM\_hidden dim  & 32    & 64      & 32    & 16      \\
		Learing\_Rate        & 5e-5  & 5e-5    & 5e-4  & 5e-5    \\ 
		Pseudo tokens num    & 4     & 4       & 4     & 4       \\
		\hline
	\end{tabular}
	\normalsize
	\caption{Hyperparameter settings for MSE-LLaMA-7B.}
	\label{table4}
	\arrayrulecolor{black}
\end{table}

\begin{figure}[ht]
	\centering
	\includegraphics[width=\linewidth,trim=0 130 580 0,clip]{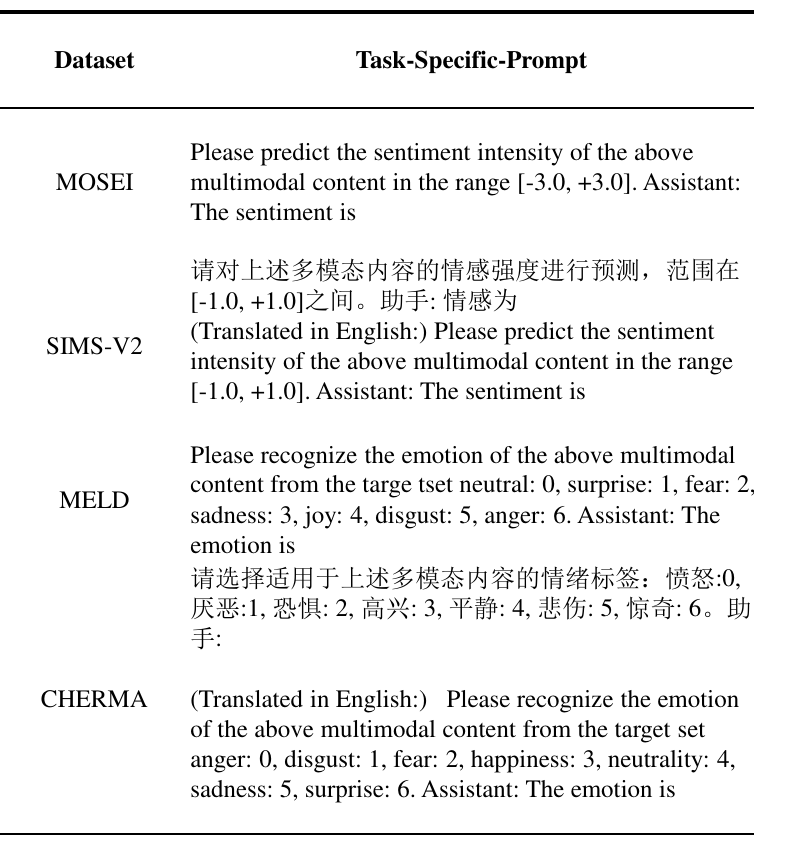}
	\caption{The Task-specific-prompt corresponding to different datasets.} 
	\label{figure 2}
\end{figure}

\begin{table*}[h]
	\centering
	\small 
	\setlength{\tabcolsep}{2.5pt} 
	\renewcommand{\arraystretch}{1.1} 
	\begin{tabular}{cccccc|ccccc|cc|cc|c} 
		\hline\hline
		\multirow{2}{*}{Model} & \multicolumn{5}{c|}{MOSEI}                                                         & \multicolumn{5}{c|}{SIMS-V2}                                                      & \multicolumn{2}{c|}{MELD}       & \multicolumn{2}{c|}{CHERMA}    & \multirow{2}{*}{\faFire Paras}  \\ 
		\cline{2-15}
		& Acc-2          & F1             & Acc-7          & MAE            & Corr           & Acc-2          & F1             & Acc2\_weak     & MAE            & Corr          & Acc            & WF1            & Acc           & WF1            &                         \\ 
		\hline
		MAG-ChatGLM3-6B        & 85.10           & 84.73          & 50.52          & 0.588          & 0.725          & 78.34          & 78.34          & 69.77          & 0.341          & 0.649         & 60.38          & 59.81          & 71.20          & 71.02          & 34.47M                  \\
		CHFN-ChatGLM3-6B       & 85.58          & 85.26          & 51.50           & 0.563          & 0.745          & 81.02          & 81.00             & 72.57          & 0.303          & 0.700           & 63.03          & 62.08          & 71.41             & 71.37              & 50.79M                  \\ 
		\hline
		MSE- ChatGLM3-6B       & \textbf{86.91} & \textbf{86.77} & \textbf{54.56} & \textbf{0.515} & \textbf{0.783} & \textbf{83.77} & \textbf{83.76} & \textbf{75.24} & \textbf{0.296} & \textbf{0.720} & \textbf{66.23} & \textbf{65.13} & \textbf{72.90} & \textbf{72.73} & \textbf{2.63M}          \\
		\hline\hline
	\end{tabular}
	\normalsize
	\caption{Experimental results on the MOSEI, SIMS-V2, MELD and CHERMA datasets. The "\faFire Paras" represents the trainable parameter of the model. }
	\label{table6}
\end{table*}

\section{APPENDIX D}
\label{appendix d}
\subsection{Baselines}
\textbf{TFN}: TFN Tensor Fusion Network (TFN) \cite{zadeh2017tensor} comprises: 1) a modality embedding sub-network that enriches encoding by taking unimodal features as inputs and outputs to a neural network, 2) a tensor fusion layer that models unimodal, bimodal, and trimodal interactions using the outer product, and 3) an emotion inference sub-network that performs emotion reasoning.

\textbf{LMF}: Low-Rank Multimodal Fusion (LMF) \cite{liu2018efficient} employs low-rank tensors to efficiently execute multimodal fusion.

\textbf{MISA}: MISA \cite{hazarika2020misa} is a multimodal framework that learns modality-invariant and modality-specific representations for each modality. The learning process is optimized through a combination of similarity loss, orthogonality loss, reconstruction loss, and task prediction loss.

\textbf{MAG-BERT}: Multimodal Adaptation Gates for BERT (MAG-BERT) \cite{rahman2020integrating} is developed by applying multimodal adaptive gates at different layers of the BERT backbone.

\textbf{Self-MM}: Self-MM \cite{yu2021learning} first utilizes a self-supervised label generation module to acquire unimodal labels, then jointly learns multimodal and unimodal representations based on multimodal labels and generated unimodal labels.

\textbf{MIMM}: MultiModal InfoMax (MMIM) \cite{han2021improving} maximizes the mutual information between unimodal inputs in pairs and between multimodal fusion results and unimodal inputs to assist the main MSA task.

\textbf{CHFN}: CHFN \cite{guo2022dynamically} is an interpretable transformer-based neural model that centres on dynamically adapting word representations to use unaligned multimodal sequences in different non-verbal contexts. It focuses on the influence of non-verbal behavioural information across the discourse spectrum and integrates this influence into the verbal representation.

\textbf{UniMSE}: UniMSE \cite{hu2022unimse} is a multimodal sentiment knowledge sharing framework that unifies MSA and ERC tasks from the aspects of features, labels, and models.

\textbf{UniSA}: UniSA \cite{li2023unisa} is a unified framework for sentiment analysis which, after pre-training and paired with task-specific prompts, can be generalized to any sentiment analysis sub-task.

\textbf{AV-MC}: The Acoustic Visual Mixup Consistent (AV-MC) \cite{liu2022make} framework utilizes unimodal annotations and unsupervised data from CH-SIMS v2.0 to learn various non-linguistic contexts for sentiment analysis.

\textbf{MMGCN}:  MMGCN \cite{hu2021mmgcn} employs Graph Convolutional Networks (GCNs) to capture contextual information, effectively leveraging multimodal dependencies and speaker information.

\textbf{MM-DFN}: MM-DFN \cite{hu2022mm} introduces a framework designed to enhance multimodal features through dynamic fusion for integrated analysis.

\textbf{EmoCaps}:  EmoCaps \cite{li2022emocaps} introduces an emotion capsule that integrates information from multiple modalities with emotional tendencies, providing a more nuanced understanding of emotions in dialogues.

\textbf{GA2MIF}: GA2MIF \cite{li2023ga2mif} presents a two-stage approach for multimodal fusion, extracting information from graphs and attention networks.

\textbf{EFT, LFT, MulT}: The early fusion transformer (EFT), Late fusion transformer (LFT), and Multi-modal Transformer (MulT) \cite{tsai2019multimodal} propose directional pairwise cross-modal attention, adapting one modality to another for multimodal fusion.

\textbf{PMR}: PMR \cite{lv2021progressive} introduces an information hub for exchanging information with each modality. This hub sends common information to each modality and reinforces their features through cross-modal attention. In turn, it also collects enhanced features from each modality, using them to generate strengthened common information.

\textbf{LFMIM}: LFMIM \cite{sun2023layer} consists of unimodal Transformer modules learning representations for each modality and a multimodal Transformer module fusing all modalities. Each module is supervised by its corresponding labels, ensuring independent learning of representations for each modality while the multimodal module aggregates all information.

\section{APPENDIX E}
\label{appendix e}
\subsection{Further discussion with models adapted for BERT}
Table \ref{table6} presents the full range of results for MAG and CHFN across the four datasets, offering further support for the discussion presented in the main text. It is noteworthy that the original MAG paper only provided MOSEI's data after sequence length alignment. Given that the implementation of MAG necessitates the utilisation of feature sequence length-aligned data, an MLP was employed to align the feature sequence lengths on the other three datasets, thereby facilitating the reproduction of the experimental results.

\section{APPENDIX F}
\label{appendix f}
\subsection{The Impact of Training Data Volume on LLM's Performance in Less Proficient Language Context}

In our prior experimental findings, we observed that LLM, even those exhibiting less proficiency in a particular language, could nonetheless demonstrate considerable capabilities on datasets of that language, provided the volume of data is sufficient. We hypothesize that this outcome is attributed to the impact of the training data volume. To further validate this impact, we conducted experiments on the MOSEI dataset using MSE-Qwen-1.8B and on the CHERMA dataset using MSE-LLaMA2-7B. For MOSEI, we reported Acc-2 and F1 score, and for CHERMA, we reported Acc and WF1 score. In this experiment, the training data was randomly sampled from the original training set in proportion, while the validation set and test set remained unchanged. Additionally, the rest of the experiment's setup was kept consistent with that in the ablation study.
\begin{figure}[!h]
	\centering
	\subfloat[MSE-Qwen-1.8B's results on MOSEI]{\includegraphics[width=0.9\linewidth]{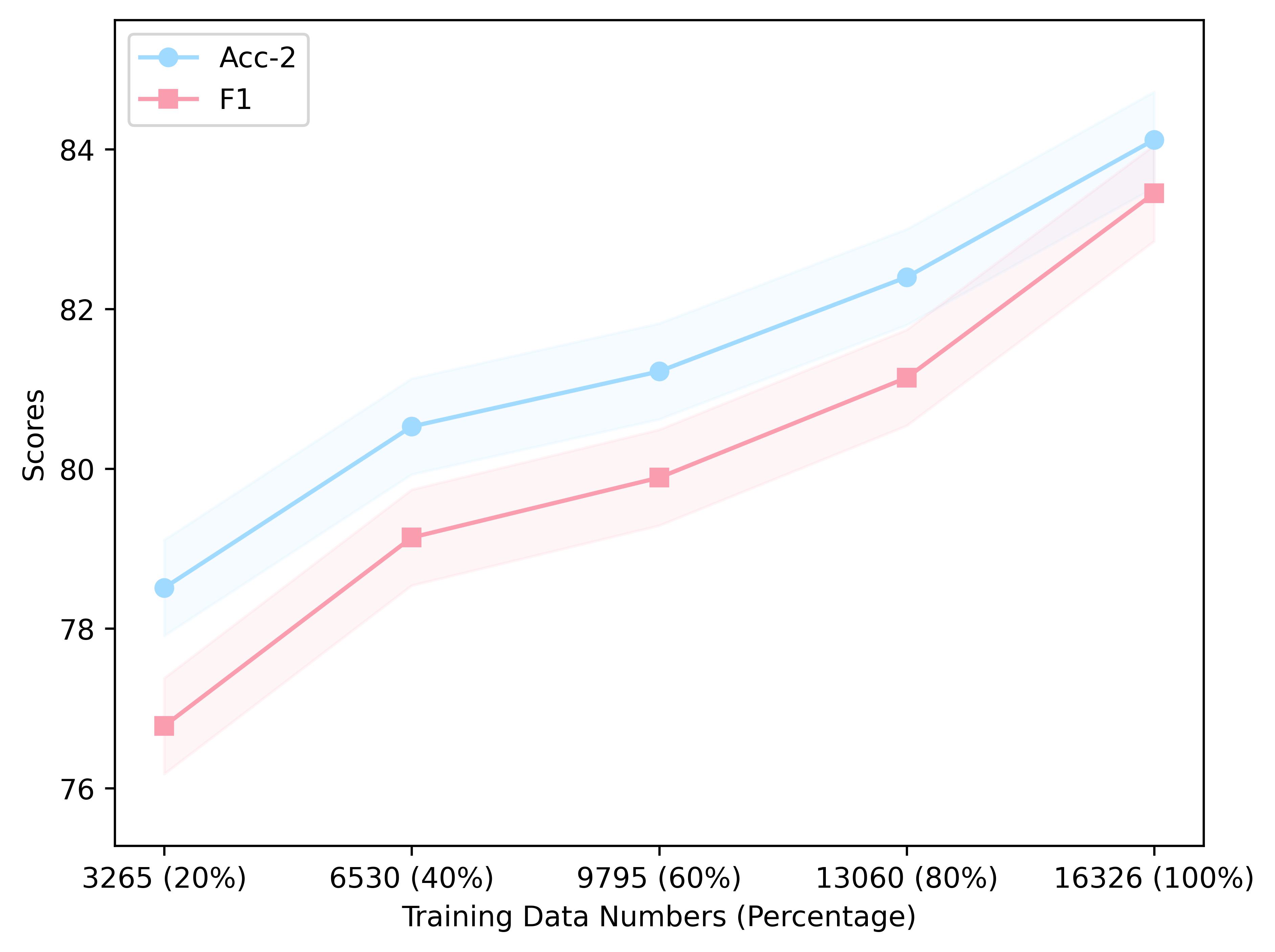}}
	\hfill
	\subfloat[MSE-LLaMA2-7B's results on CHERMA]{\includegraphics[width=0.9\linewidth]{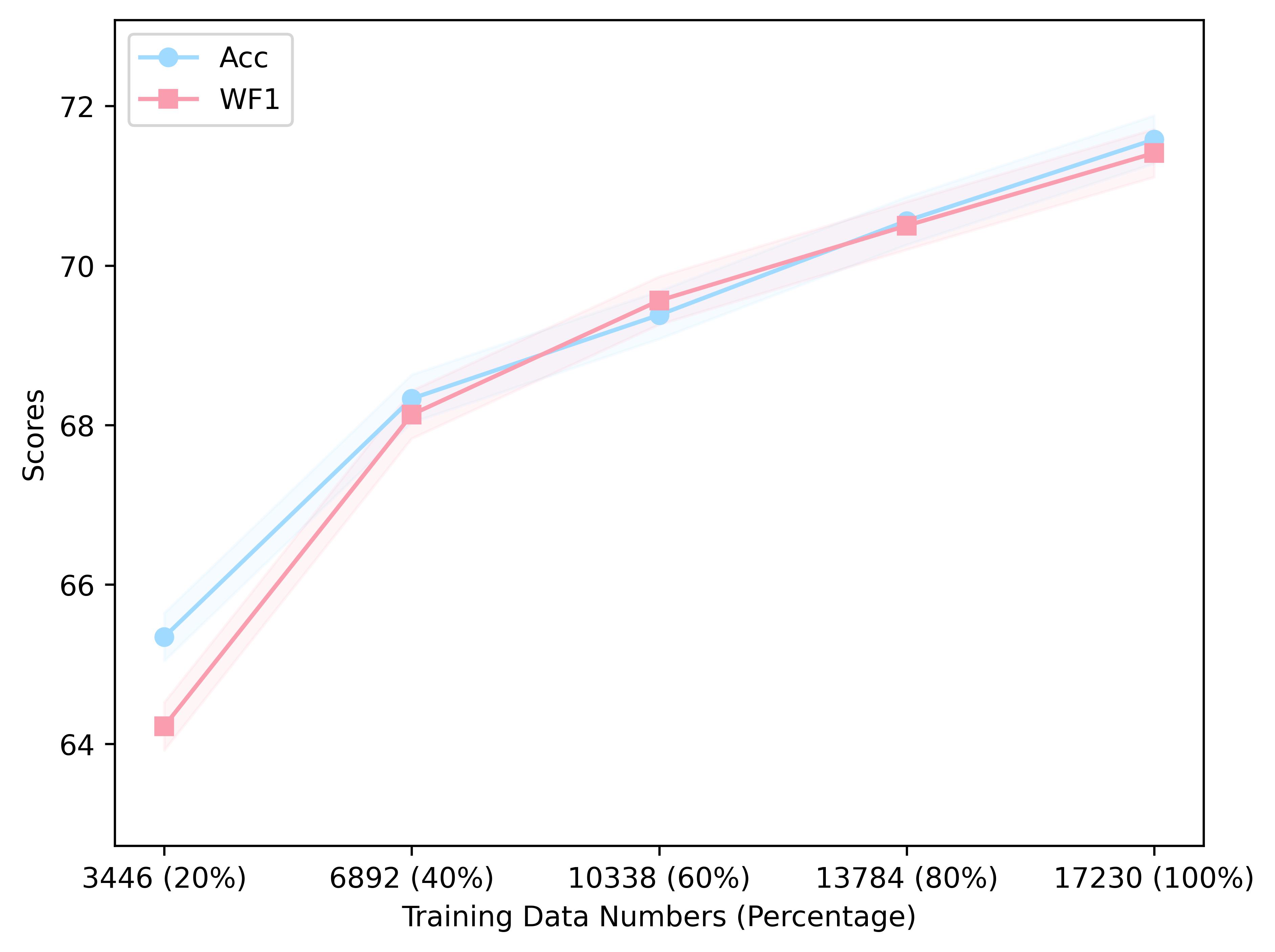}}
	\caption{Performance of MSE-Adapter with different number of training data.}
	\label{figure 3}
\end{figure}

The results of the experiment are shown in Figure \ref{figure 3}. The figure indicates a significant decrease in the performance of MSE-Qwen-1.8B on MOSEI when the quantity of training data dropped below 40\% of the training set (approximately 6530 samples). Similarly, it was found that MSE-LLaMA2-7B also experienced a decrease in performance on CHERMA when the training samples were less than 40\% of the total training set size (approximately 6892 samples). Furthermore, when trained with datasets comprising over 10,000 instances, these models show decent performance. Therefore, it can be argued that MSE-Qwen-1.8B/MSE-LLaMA2-7B can still show commendable performance when the training data exceeds 10,000 instances, despite its slightly inferior English/Chinese proficiency. In summary, it can be concluded that the size of the training data has a certain impact of training data volume on LLM's performance in less proficient language context.

\end{document}